%% file: main.tex
\documentclass[conference]{IEEEtran}
\usepackage{times}

\usepackage[numbers]{natbib}
\usepackage{multicol}
\usepackage[bookmarks=true]{hyperref}
\usepackage{graphics} 
\usepackage{epsfig} 
\usepackage{mathptmx} 
\usepackage{times} 
\usepackage{amsmath} 
\usepackage{amssymb}  
\usepackage{wrapfig}

\usepackage{subfigure}
\usepackage{graphicx}
\usepackage{algorithm}
\usepackage{algcompatible}
\usepackage{wrapfig}
\usepackage{xcolor}
\usepackage{gensymb}
\usepackage{hyperref}

\newcommand*{\method}{DBAP}

\DeclareMathOperator*{\argmax}{arg\,max}

\begin{document}

\title{Demonstration-Bootstrapped Autonomous Practicing via Multi-Task Reinforcement Learning}




%
\author{\authorblockN{Abhishek Gupta\authorrefmark{1},
Corey Lynch\authorrefmark{2},
Brandon Kinman\authorrefmark{2}, 
Garrett Peake\authorrefmark{2},
Sergey Levine\authorrefmark{1} and Karol Hausman\authorrefmark{2}}
\authorblockA{\authorrefmark{1} UC Berkeley}
\authorblockA{\authorrefmark{2} Google Inc}}

\maketitle

\begin{abstract}
Reinforcement learning systems have the potential to enable continuous improvement in unstructured environments, leveraging data collected autonomously. However, in practice these systems require significant amounts of instrumentation or human intervention to learn in the real world. In this work, we propose a system for reinforcement learning that leverages multi-task reinforcement learning bootstrapped with prior data to enable continuous autonomous practicing, minimizing the number of resets needed while being able to learn temporally extended behaviors. We show how appropriately provided prior data can help bootstrap both low-level multi-task policies and strategies for sequencing these tasks one after another to enable learning with minimal resets. This mechanism enables our robotic system to practice with minimal human intervention at training time, while being able to solve long horizon tasks at test time. We show the efficacy of the proposed system on a challenging kitchen manipulation task both in simulation and the real world, demonstrating the ability to practice autonomously in order to solve temporally extended problems. 
\end{abstract}

\IEEEpeerreviewmaketitle

\input{texs/introduction}
\input{texs/related_work}
\input{texs/preliminaries}

\input{texs/method}
\input{texs/system_description}
\input{texs/experiments}
\input{texs/discussion}
\input{texs/acknowledgements}

\bibliographystyle{plainnat}
\bibliography{example}

\appendix 

\input{texs/appendix}

\end{document}

%% file: texs/introduction.tex
\section{Introduction}
\label{sec:intro}

\begin{figure}[!t]
    \centering
    \includegraphics[width=0.9\linewidth]{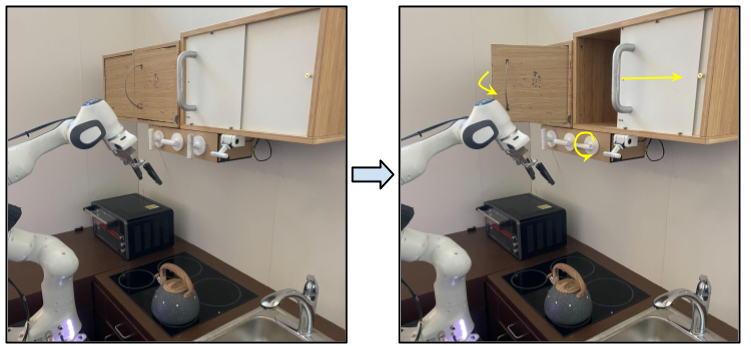}
    \caption{\footnotesize{Robot setup for real world training in a kitchen. The robot needs to manipulate multiple different elements to accomplish complex goals.}}
    \label{fig:robot_env}
    \vspace{-0.6cm}
\end{figure}

Consider a robot deployed in a previously unseen kitchen, such as the one shown in Fig.~\ref{fig:robot_env}. Since this is an unfamiliar environment, the robot might not know exactly how to operate all the kitchen appliances, open cabinets, operate the microwave or turn on the burners, especially if these need to be done in a specific sequence resulting in a long-horizon task. Like many intelligent agents, the robot needs some practice, making reinforcement learning (RL) a promising paradigm. However, as we have learn from prior work, practicing long-horizon tasks with RL is difficult for reasons such as exploration~\cite{gupta2019relay}, environment resets~\cite{han2015learning}, and credit assignment problems~\cite{harutyunyan2019hca}. The process of building robotic learning systems (especially with RL) is never as simple as placing a robot in the environment and having it continually improve. Instead, human intervention is often needed in the training process, especially to provide resets to the learning agent. Ideally, we would want to provide the robot with a small number of examples to help with these challenges in the training process and do so at the beginning of the practicing process so that it can continually practice a wide range of tasks in the kitchen on its own, with minimal human intervention. 

In this work, we consider the following perspective on this problem. Rather than only providing examples that aim to aid the final solution (e.g. by providing demonstrations of the final long-horizon tasks), we additionally focus on providing supervision that helps the robot to continue practicing long-horizon tasks autonomously. 
In particular, we provide a set of continuous, uninterrupted multi-task demonstrations~\cite{lynch2020learning, gupta2019relay} that illustrate example behaviors and goal states that the robot should be able to acquire and robustly achieve. The continuous aspect of the demonstrations provides two benefits: i) it breaks up the long-horizon tasks into smaller subgoals that can be used as intermediate points, and ii) it provides data \textit{in-between} the tasks, showing the robot examples of behaviors where some of the tasks are not fully finished but are reset using other tasks - a case that often happens during the autonomous execution of the policy.
Equipped with the continuous demonstrations, we can learn how to sequence different tasks one after the other and, use that ability to have some tasks provide resets for others, enabling the robot to practice many tasks autonomously by resetting itself without actually needing constant human intervention in the learning process. For instance, if the robot is practicing opening and closing the cabinet in Fig.~\ref{fig:robot_env}, it may start in a state where the cabinet is closed and command the task to open it. Even if this task does not succeed and the cabinet is partially open, the agent can continue practicing any of the remaining tasks including closing or opening the cabinet to minimize the requirement for resets, naturally resetting the environment by just solving different tasks. 
At test time, we can then utilize the same ability of sequencing different tasks to sequence multiple subgoals for completing long-horizon multi-stage tasks.

In this paper, we present a novel robotic learning system, which we call demonstration bootstrapped autonomous practicing (DBAP), that is able to incorporate a small amount of human data provided upfront to bootstrap long periods of autonomous learning with just a few human interventions for episodic resets, using some tasks to reset others while learning how to perform long horizon behaviors. We show that just a few hours of easily-provided ``play-style''~\cite{lynch2020learning, gupta2019relay} multi-task demonstrations can make the entire learning process more efficient, bootstrapping both policy learning and autonomous practicing of behaviors, eventually showing the ability to perform long-horizon multi-step tasks. Our experiments indicate that this paradigm allows us to efficiently solve complex multi-task robotics problems both in simulation and in a real world kitchen environment, with an order of magnitude less human intervention needed during learning than typical RL algorithms.

%% file: texs/related_work.tex
\section{Related Work}
\label{sec:related}

Most prior applications of RL in robotic manipulation rely on an episodic reset mechanism, which brings the robot back to a static distribution over initial states. Recent work has attempted to relax the episodic reset assumption~\cite{co2020ecological,zhu2020ingredients, eysenbach2017leave, han2015learning} by either learning an explicit reset controller~\cite{eysenbach2017leave, han2015learning} or learning a perturbation controller~\cite{zhu2020ingredients} to randomly reset the state. As opposed to these works, we operate in the continual multi-task setting. Recent work~\cite{gupta2021reset, ha2020learning, han2015learning} has approached the problem through the lens of multi-task learning, but require significant human effort to design how tasks should be sequenced, relying on human-defined state machines. In contrast, our work shows how we can bootstrap the autonomous learning process with human-provided prior data, obviating the need for hand definition of task sequencing state-machines. 

There are multiple approaches that aim to use demonstrations to provide a good initialization for policies via learning from demonstrations~\cite{mulling2013learning, manschitz2015learning, rajeswaran2017learning, gao2018reinforcement, billard2008survey, mandlekar2020gti}. In a similar vein, algorithms for offline RL \cite{levine2020offline} aim to use large datasets to bootstrap learning, without making explicit assumptions about the provided data. Most algorithms for offline RL aim to enable training from static offline datasets, by preventing the off-policy bootstrap from diverging due to internal distribution shift~\cite{levine2020offline, fujimoto2019off, kumar2019stabilizing,  mandlekar2020iris, kumar2020conservative, peng2019advantage, wang2020critic, nair2020accelerating, kostrikov2021offline, singh2020cog, yu2021combo}. While most of these are in purely offline settings, recently there have been approaches~\cite{nair2020accelerating,deploymentorl} that bootstrap initial policies offline but aim to also continue finetuning. In contrast to these papers, which utilize offline data simply as an initialization for a task policy that can be finetuned online but assume complete access to \emph{episodic resets}, our aim is to instead utilize the provided data to learn how to perform \emph{autonomous practicing}: for our method the prior data doesn't just provide a good initialization for a single task, but a set of useful behaviors that the agent can use to reset itself after failed attempts, and chain together to perform multi-stage behaviors. 

Long horizon problems have often been approached from the angle of hierarchical RL algorithms ~\cite{dayan92feudal, dietterich98maxq, sutton99options, plb2017optioncritic, nachum18hiro, gupta2019relay, levy2019hac, eysenbach2019sorb, savinov18sptm, kreidieh2019marlhrl, barto2003hrl}. These algorithms aim to learn high-level policies that sequence low-level policies to solve temporally extended tasks. As opposed to these methods, in addition to sequencing low-level tasks for long-horizon goals, our system is also used as a means to perform autonomous practicing with minimal human interventions.  Perhaps most closely related to our work, relay policy learning (RPL) ~\cite{gupta2019relay}, uses demonstrations to bootstrap a hierarchical learning setup. However, RPL does not address the question of human effort and assumes episodic resets at training time, which DBAP alleviates. 

%% file: texs/preliminaries.tex
\section{Preliminaries and Problem Statement}
\label{sec:prelim}


We use the standard MDP formalism $\mathcal{M} = (S, A, P, r, T)$ (state, action, dynamics, reward, episodic horizon), but extend RL to goal-reaching tasks, where the task is represented as a state goal $s_g$ that the goal-conditioned policy $\pi_\theta(a|s, s_g)$ has to reach to obtain the reward. We assume to be given a set of $N$ goal states that are of particular interest: $S_{g_i} = \{s_g^i\}_{i=1}^N$ that the agent is expected to to reach from any other state of interest, resulting in the following objective: 
\vspace{-4pt}
\begin{equation}
	\label{eq:obj1}
	\max_{\theta} \mathbb{E}_{s_0 = s_g^i, a_t \sim \pi_{\theta}(a_t|s_t, s_g^j)}\sum_{t=1}^T r(s_t, a_t, s_g^j) \hspace{0.5cm} \forall s_g^i \in S_{g_i}, s_g^j \in S_{g_i}
\end{equation}




We often refer to reaching these goals of interests as \textit{tasks} since reaching such states usually corresponds to accomplishing a meaningful task such as opening a cabinet. Note that reaching certain $s_g^j$ from $s_g^i$ would involve just short horizon behaviors spanning a single episode, while reaching other $s_g^j$ from $s_g^i$ may require long horizon reasoning over multiple different sub-steps, for instance it may require reaching $s_g^k, s_g^l, ...$ on the path from $s_g^i$ to $s_g^j$.

In order to allow for more autonomous learning with minimal human supervision, we assume that a reset can be provided every $n$ episodes (or every $nT$ time steps). 
Our objective in this case is to train the agent in a training MDP $\mathcal{M}' = (S, A, P, r, nT)$, where resets are only received every $nT$ steps such that we are able to learn a policy $\pi$ that is effective on the original episodic problem described in Eq.~\ref{eq:obj1}. This requires our learning algorithm to operate in a non-stationary setting and learn how to practice tasks mostly autonomously with only occasional resets as discussed in several prior works ~\cite{eysenbach2017leave, gupta2021reset, zhu2020ingredients}.

\textbf{AWAC:} 
As we will discuss next, solving this problem effectively for long horizon tasks requires us to incorporate demonstrations to bootstrap both the policy and the autonomous practicing mechanism. In order to effectively incorporate prior demonstrations to bootstrap policy learning, we use the recently developed AWAC algorithm~\cite{nair2020accelerating}. The main idea behind AWAC is to constrain the actor policy to be within the training data distribution, which addresses the problem of overly optimistic Q-function estimates on unseen state-action pairs~\cite{kumar2019stabilizing, kumar2020conservative}.
In particular, the AWAC algorithm applies the following weighted maximum likelihood update to the actor:
$\theta_{k+1} = \argmax_\theta \mathbb{E}_{s, a} \left[\log \pi_\theta(a|s) \exp \left(\frac{1}{\lambda}A^{\pi_\theta}(s, a) \right)\right]$,
where is $A$ is the advantage function, and $\lambda$ is a hyper-parameter. Since the actor update in AWAC samples and re-weights the actions from the previous policy, it implicitly constrains the policy. We directly build on this framework for policy bootstrapping but in the following section show how we can additionally bootstrap autonomous practicing from offline data as well.

%% file: texs/method.tex
\begin{figure*}[!t]
    \centering
    \includegraphics[width=\textwidth]{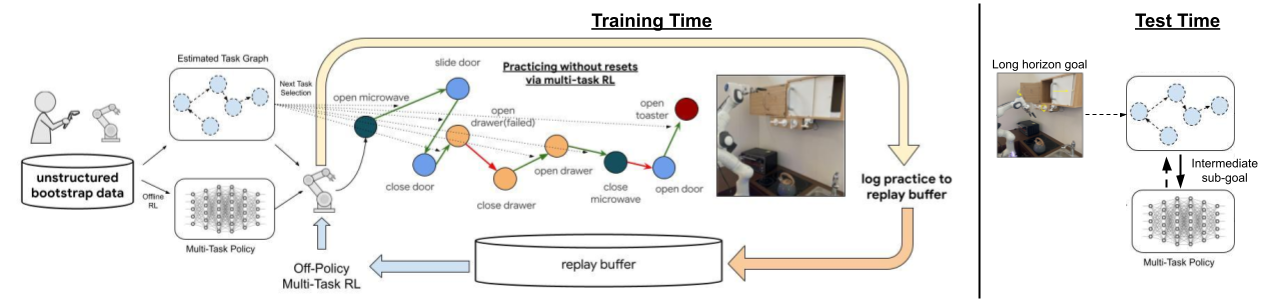}
    \caption{\footnotesize{Given human-provided unstructured demonstrations, the system bootstraps a multi-task RL policy via offline RL and builds a task graph that models transitions between different tasks. The system then practices the tasks autonomously with small number of resets, using the task graph to command the appropriate next task. The resulting multi-task policy and task graph are then used to solve long-horizon problems at test-time.}}
    \label{fig:mainfig}
    \vspace{-0.65cm}
\end{figure*}

\section{Demonstration Bootstrapped Autonomous Practicing for Multi-Task Reinforcement Learning}
\label{sec:method}
Our method, Demonstration-Bootstrapped Autonomous Practicing (DBAP) learns how to perform complex behaviors in an environment by leveraging human-provided demonstration data in two ways: to bootstrap a multi-task policy that learns to perform a variety of behaviors; and to enable a graph-based sequencing mechanism, which allows for autonomous improvement by choosing what sequence of these behaviors to command to minimize number of resets at training time. In doing so, this sequencing mechanism  helps to practice each behavior at training time while also enabling the robot to sequence practiced behaviors to accomplish long-horizon tasks at test time. An overview of our overall framework is presented in Fig.~\ref{fig:mainfig}. Given a human-provided dataset, we leverage an offline RL algorithm to bootstrap a multi-task policy, while using a graph search algorithm to i) decide how to sequence tasks for practicing to improve this multi-task policy with minimal resets at training time and ii) deciding how to sequence behaviors to solve long-horizon tasks at test-time.




More precisely, we learn a single goal-conditioned policy for multiple tasks, denoted $\pi_\theta(a|s, s_g)$, where $s_g$ is a goal state. To learn this goal-conditioned policy, we instantiate a goal-conditioned version of AWAC~\cite{nair2020accelerating}, where the policy is provided with the state goal $s_g$ alongside the current state $s$, yielding the following policy update:
$\theta_{k+1} = \argmax_\theta \mathbb{E}_{s, a, s_g} \left[\log \pi_\theta(a|s, s_g) \exp \left(\frac{1}{\lambda}A^{\pi_\theta}(s, a, s_g) \right)\right]$, 
where $A^{\pi_\theta}$ is an advantage function for the current, goal-conditioned policy $\pi_\theta$. This goal conditioned policy is directly bootstrapped from human provided data as described in prior work ~\cite{nair2020accelerating} Prior data is not just used for policy bootstrapping but to autonomously improve this bootstrapped multi-task policy through autonomous practicing with minimal numbers of human resets, while also being able to leverage the policy and practicing mechanism appropriately to solve long horizon problems. 


In order to enable agents to practice on their own with minimal number of human interventions, we require a task-sequencing policy that determines how the goals are sequenced at training time.
In particular, a task-sequencing policy $\pi^\text{hi}(s_g|s, s_g^{\text{desired}})$ decides which goal of interest $s_g$ to command next from the current state during practicing. 
This resembles high-level policies used in hierarchical RL that are commonly used to sequence individual subgoals for accomplishing long-horizon goals. 
However, in this scenario, $\pi^\text{hi}(s_g|s, s_g^{\text{desired}})$ is not just used to accomplish long-horizon goals at test-time but also to enable autonomous practicing at training time. In the following sections we will often refer to the multi-task policy $\pi_\theta(a|s, s_g)$ as the low-level policy, and the task sequencer $\pi^\text{hi}(s_g|s, s_g^{\text{desired}})$ as the high-level policy. 


\textbf{Assumptions:} Before we dive into system details, we discuss some assumptions made in our problem setting. Firstly, we assume that the environment does not end up in irrecoverable states from which no action can bring the environment back to goals of interest. This assumption is common across a variety of reset-free RL algorithms ~\cite{gupta2021reset, han2015learning, zhu2020ingredients, eysenbach2017leave, ha2020learning} and spans a wide range of practical tasks in robotic manipulation and locomotion.
We also assume that the states can be discretely grouped into different tasks or goals of interest, which in our case is done by a discrete thresholding operation (more details on our supplementary website \mbox{\url{https://dbap-rl.github.io/}}. This can be extended to high-dimensional environments using clustering or representation learning algorithms.

\begin{figure}[!h]
\vspace{-0.4cm}
\begin{minipage}{.45\textwidth}
\begin{algorithm}[H]
\caption{PracticeGoalSelect: High Level Task Selection via Graph Search}
\label{algo:taskselection}
\textbf{Input:} task-graph $\mathcal{G}$, goal-states $\{s_g^0, .., s_g^N\}$, density over goals $\rho(s_g^i)$, current state $s$

\textbf{Output:} Next goal $s_g^{\text{next}}$ to command
\begin{algorithmic}
    \STATE \textcolor{blue}{// Initialize maximum entropy value}
    \STATE $\mathcal{H}_{\text{max}} = -\infty$  
\FOR{$s_g^i \in \{s_g^1, \dots, s_g^N\}$} 
    \STATE $\tau \leftarrow \text{Dijkstra}(s, s_g^i, \mathcal{G})$ 
    \STATE $\rho' = \rho + \tau$ \textcolor{blue}{// Compute updated density}
    \IF{$\mathcal{H}(\rho') \geq \mathcal{H}_{\text{max}}$}
        \STATE  $\mathcal{H}_{\text{max}} = \mathcal{H}(\rho')$
        \STATE  $s_g^{\text{next}} \leftarrow \tau[1]$ \textcolor{blue} {// First step of path to $s_g^i$}
    \ENDIF
\ENDFOR
\STATE{\textbf{return}  $s_g^{\text{next}}$}
\end{algorithmic}
\end{algorithm}
\end{minipage}
\hspace{0.2cm}
\begin{minipage}{.45\textwidth}
\begin{algorithm}[H]
\caption{Overview of \method}
\label{algo:system_overview}
\textbf{Input:} Human provided multi-task demonstrations $\mathcal{D}$

\textbf{Output:} Multi-task policy $\pi_{\theta}(a|s, s_g)$ and task-graph $\mathcal{G}$
\begin{algorithmic}
\STATE Estimate task graph $\mathcal{G}$ via counting
\STATE Initialize $\pi_{\theta}(a|s, s_g)$ via AWAC~\cite{nair2020accelerating}
\STATE Initialize current density $\rho = 0$
\FOR{t = 0, \dots, $N$ steps} 
    \STATE \textcolor{blue} {// Select next goal via graph search}
    \STATE $s_g^{\text{next}} \leftarrow \mathrm{PracticeGoalSelect}(s, \rho)$
    \STATE Rollout $\pi_{\theta}(a|s, s_g^{\text{next}})$ 
    \STATE Add collected data to replay buffer $\beta$ 
    \STATE Update $\pi_{\theta}$ via AWAC. 
\ENDFOR 
\STATE{\textbf{return} $\pi_{\theta}(a|s, s_g)$, $\mathcal{G}$}
\end{algorithmic}
\end{algorithm}
\end{minipage}
\vspace{-0.5cm}
\end{figure}

\subsection{Task Sequencing via Graph Search}
\label{sec:method-gs} 

A successful high-level task sequencer policy $\pi^\text{hi}$ is one that at training time is able to propose goals for autonomous practicing, while at test time is able to direct the agent towards reaching long-horizon goals. To learn such a task sequencer policy $\pi^\text{hi}$, we propose a simple model-based graph search algorithm. Note that $\pi^\text{hi}$ is not a neural network, but the result of a search procedure. As we show empirically in our experimental evaluation, this is more flexible and robust than parameterizing $\pi^\text{hi}$ directly with a neural network. 

The key idea in our approach is to leverage prior data to learn which low-level task transitions are \emph{possible} and can be sequenced, and then use this knowledge to optimize for \emph{both} autonomous practicing (at training time) and long-horizon goal sequencing (at test time). 
In particular, we utilize the provided data to build a directed task graph $\mathcal{G}$, with vertices as different goal states of interest $s_g^i$, $s_g^j$, and an adjacency matrix $A$ with $A(i, j) = 1$ if there exists a transition between particular states of interest $s_g^i$ and $s_g^j$ in the demonstration data, and $A(i, j) = 0$ otherwise. This graph can be thought of as a discrete high-level model, which represents how different goal states of interest are connected to each other. Given this graph $G$ acquired from the prior data, we can then utilize it for both autonomous practicing of low-level tasks and for sequencing multiple low-level tasks to achieve multi-step goals. We describe each of these phases next.

\textbf{Autonomous practicing (training time).} As described in prior work ~\cite{zhu2020ingredients}, maintaining a uniform distribution over tasks helps when learning with minimal resets by ensuring all tasks are equally represented in the learning process. The task graph $\mathcal{G}$ is used to direct autonomous practicing by explicitly optimizing to command goals that bring the overall coverage over states close to the uniform distribution, while also minimizing the number of human interventions.

In particular, the algorithm iterates through all possible goal states of interest $s_g^i$, determines the shortest path from the current state to the goal state via Dijkstra's algorithm~\cite{dijkstra1959note}, and computes the resulting densities over goal states if that path was chosen. The path that results in bringing the density closest to uniform (i.e. maximum entropy) is then picked, and the first step in the path (i.e. a short-horizon goal) is commanded as the next goal state. 
This ensures that the algorithm maintains uniform coverage over different goal states of interest in the environment, so that all tasks can be practiced equally. This process repeats at the next step, thereby performing receding horizon control~\cite{garcia89surveympc} to maintain the density over potential goal states of interest to be as close to uniform as possible, while training the policy to keep practicing how to accomplish various short horizon goals. 

Formally, the objective being optimized to select which goal to sequence next is given by: \mbox{$\max_{s_g^i \in S_g^i} \mathcal{H}(\mathcal{U}, \rho + \text{Djikstra}(s, s_g^i))$},
where $\rho$ is the current marginal distribution (represented as a categorical distribution) over goal states of interest, $\text{Dijkstra}(s, s_g^i)$ computes the shortest distances between current state $s$ and $s_g^i$ and the goal is to bring the updated density as closer to uniform $\mathcal{U}$\footnote{We overload the $+$ operator here to denote an update to the density $\rho$ when accounting for new states.}. A detailed description can be found in Algorithm~\ref{algo:taskselection}.

\textbf{Task sequencing for long-horizon tasks (test time).} Next, we ask how short-term behaviors learned by the multi-task policy $\pi_\theta(a|s, s_g)$ can be sequenced to accomplish long-horizon goals. We note that the same graph search mechanism used for autonomous practicing at training time can be reused to sequence appropriate sub-goals at test time to accomplish long horizon tasks.

More formally, given a target goal of interest $s_g^j$ when the agent is at a particular state $s$, we can use the estimated graph $\mathcal{G}$ via Dijkstra's algorithm to compute the shortest path between the current state and the goal of interest $s_g^j$ ($\tau = \left[s, s_g^1, s_g^2, \dots, s_g^j\right]$). The next goal state of interest in the path $s_g^1$ is then chosen as the next goal commanded to the multi task policy $\pi(a|s, s_g^1)$ and executed for a single episode of $T$ steps till the agent reaches $s_1$. This procedure is repeated until the agent accomplishes $s_g^j$. Further details on these procedures can be found in Algorithm~\ref{algo:taskselection} and Algorithm~\ref{algo:system_overview}. 

%% file: texs/system_description.tex
\section{System Description}
\label{sec:system}


To evaluate (DBAP) in the real world, we built a physical kitchen environment based on the kitchen manipulation benchmark described in prior work \cite{gupta2019relay}.  


\begin{figure*}[!h]
    \centering
    \includegraphics[width=\linewidth]{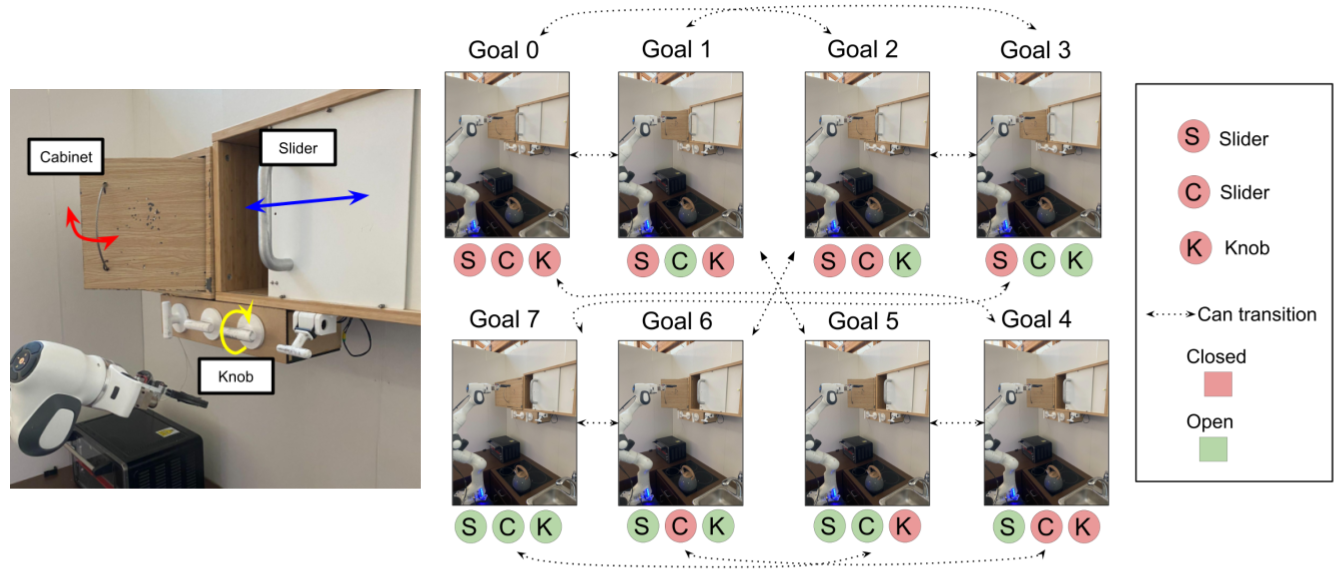}
    \caption{\footnotesize{Elements, tasks, and goals in the real-world kitchen environment. The agent is manipulating the cabinet, slider and knob to accomplish particular configurations as shown in goals $0$ to $7$. The dotted lines represent individual transitions, toggling one element at a time between its extreme positions. The goal of the agent is to learn a policy and a graph controller that is able to transition between goal states.}}
    \label{fig:real_tasks}
    \vspace{-0.2cm}
\end{figure*}



\textbf{Tasks.} In this environment, we task a 7 DoF Franka Emika robotic arm with manipulating three distinct elements: a sliding door that can be slid open or closed, a cabinet door that can be open or closed and a knob that can rotate to control the stove burners, as shown in Fig~\ref{fig:real_tasks}. At each time step, the policy chooses a position and orientation for the end effector, or opens/closes the gripper.
These three elements represent distinct types of motion, each of which require different control strategies. The goal states of interest $s_g^i$ are defined as combinations of the elements being opened or closed (or in the case of the knob, turned by $0\degree$ or $90\degree$), resulting in $2^3 = 8$ goal states based on combinations of the elements being open or closed (Fig~\ref{fig:real_tasks}). 
As described in Sections~\ref{sec:prelim}, ~\ref{sec:method}, the agent must practice reaching the goals of interest autonomously in the environment to master going from any combination of element configurations to any other. The agent is said to be in one of the goal states of interest if the current state of the elements in the scene are within $\epsilon$ distance of the goal. Long horizon goals involve flipping the state (from open to close or vice versa) of all three elements of the scene, requiring sequential reasoning.

\begin{figure}[!h]
    \centering
     \includegraphics[width=0.4\textwidth]{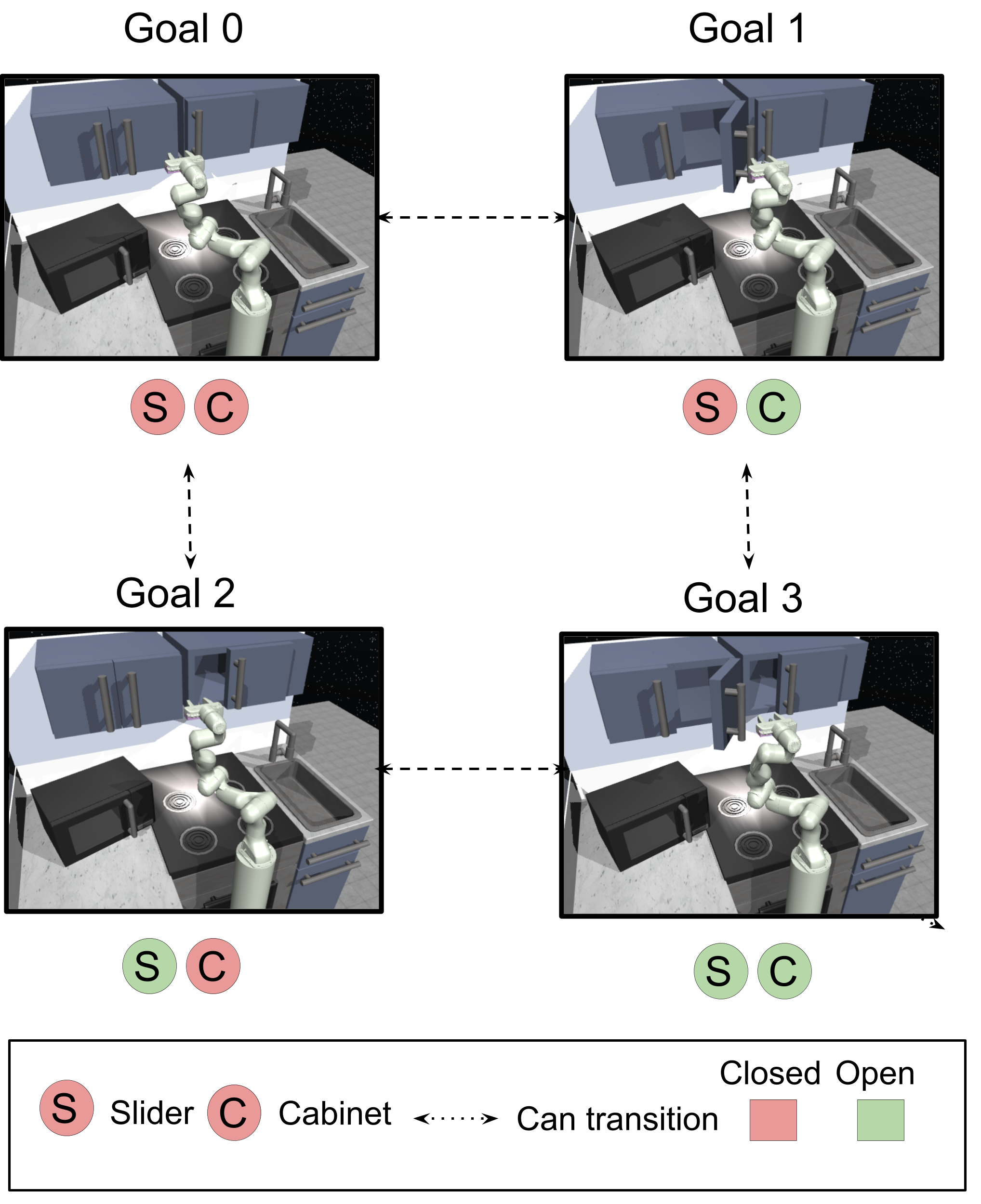}
    \caption{\footnotesize{Simulated tasks in kitchen. Tasks involve manipulating the kitchen cabinet and the sliding door to achieve various combinations of configurations.}}
    \label{fig:sim_env}
\end{figure}

\textbf{Data collection.} We make use of a teleoperation system to provide a continuous sequence of multi-task demonstrations. We collect ``play-style''~\cite{lynch2020learning, gupta2019relay} demonstrations, where different tasks are attempted successfully one after the other, indicating both how tasks are solved and how they may be sequenced. We make a simple change to the data collection procedure where the user indicates when a particular goal state of interest is completed before transitioning over to demonstrating a different goal. This allows the algorithm to easily determine the goals of interest as the transition points between these human-provided demonstrations. We provide around $500$ demonstrations in the real world, requiring $2.5$ hours of data collection time. 

\subsection{Simulation Environment}

To provide thorough quantitative comparisons, we also evaluate our method on a simulated version of the above task, based on the MuJoCo kitchen manipulation environment described out by \cite{gupta2019relay}. The purpose of this evaluation is to study the behavior of different components of the system in detail and more systematically run comparisons. In particular, in simulation we consider tasks with 2 elements (Fig~\ref{fig:sim_env}): the cabinet and the slider. The  goal states correspond to combinations of cabinet and slider being open and closed. 

%% file: texs/experiments.tex
\section{Experimental Evaluation}
In our experimental evaluation, we aim to answer the following questions: 
\textbf{(1)} Does DBAP allow agents to learn how to perform long horizon behaviors in the real world?
\textbf{(2)} Does DBAP reduce the number of human resets required to learn long horizon behaviors, as compared to existing methods?
\textbf{(3)} Is DBAP able to use prior data to bootstrap the learning of both low-level policy execution and high level practicing behavior?



To understand the decisions made in DBAP, we compare to a number of baselines in simulation and the real world. For real-world experiments, we compare DBAP to behavior cloning without any finetuning (``Imitation" in Table~\ref{tab:success-rates-goalreaching-realworld}), offline RL with AWAC without finetuning (``Offline RL" in  Table~\ref{tab:success-rates-goalreaching-realworld}) and the no pre-training baseline. The policies operate on a low-level state of object and end effector positions, and they are represented with 3 layer MLPs of $256$ units each. Find more details of the method, system  architecture, ablations, visualizations of the data at our supplementary website \mbox{\url{https://dbap-rl.github.io/}} as well as a short video overview of the method and videos of learned behaviors.

In simulation we compared with the following: \textbf{i) Non-pretrained low level, graph search high level:} This is a version of our algorithm, where the human-provided data is not used to bootstrap low level policies, but only the high level graph. This is related to ideas from ~\cite{gupta2021reset}, where multiple policies are learned from scratch, but unlike that work this baseline uses a learned high level graph. \textbf{ii) Pretrained low level, random high level controller:} This is a multi-task analogue to the perturbation controller~\cite{zhu2020ingredients}. The low-level policy is pre-trained from human demonstrations, but the high-level practicing chooses randomly which task to execute. 
\textbf{iii) Pretrained low level, BC task selection:} This is a version of the method by~\cite{gupta2019relay}, modified for the minimal reset setting. Rather than using random task selection for practicing, tasks are sequenced using a high level policy trained with supervised learning on the collected data.
\textbf{iv) Pretrained low level, reset controller:} 
This baseline is similar to a reset controller~\cite{han2015learning, eysenbach2017leave}, and alternates between commanding a random task and commanding a return to a single fixed start state during practicing. In this case, the high-level policy is a random controller.
\textbf{v) Imitation low level, imitation high level:} This baseline involves training the low level and high level policies purely with imitation learning and running an offline evaluation of how well the policy performs (as in ~\cite{le2018hirl}).
\textbf{vi) Full relabeled relay policy learning (Gupta et al.):} It involves training the RPL algorithm~\cite{gupta2019relay} with fully relabeled goals, as opposed to just using labeled changepoints.

\textbf{Evaluation metrics.}
We evaluate over multiple long horizon goal reaching manipulation tasks in the real world environment shown in Fig~\ref{fig:real_tasks} and simulation environment shown in Fig~\ref{fig:sim_env}. Our real world environment has three elements (cabinet, knob, and slider) and we consider two possible goal states per element (fully open or fully closed). Each long horizon task requires changing the original state of the environment to the inverted goal state. Success requires that a given task sequencer $\pi^\text{hi}$ chain multiple low-level goal reaching behaviors to accomplish the long-horizon goal. For example, given original state ``cabinet open, knob open, and slider closed", the goal state would be ```cabinet closed, knob closed, and slider open". We construct 8 long horizon tasks in this manner and report the average success rate.

\begin{table}[!hb]
        \centering
        \begin{tabular}{|l|l|l|} \hline
        & \textbf{Success Rate}              & \textbf{Path Length}\\ \hline
        Offline RL & $0.83 \pm  0.058$               & $ 3.5 \pm  0.17$             \\ \hline
    
      \textbf{DBAP (Ours)}  & $ \mathbf{0.95 \pm 0.05}$              & $\mathbf{3.37
     \pm  0.3}$              \\ \hline
        Imitation & $ 0.62 \pm 0.$ & $4.3 \pm 0.15$ 
     \\ \hline
        No Pretraining & $0.0 \pm 0.0$ & $6.0 \pm 0.0$  \\ \hline
        \end{tabular}
        \caption{\footnotesize{Success rates and path lengths (in number of steps) in the real world for reaching long horizon goals.}}
        \label{tab:success-rates-goalreaching-realworld}
    
    \vspace{-1cm}
\end{table}

\subsection{Demonstration-Bootstrapped Practicing in Real World}
\begin{figure*}[!h]
\vspace{-0.4cm}
    \centering
    \includegraphics[width=0.8\textwidth]{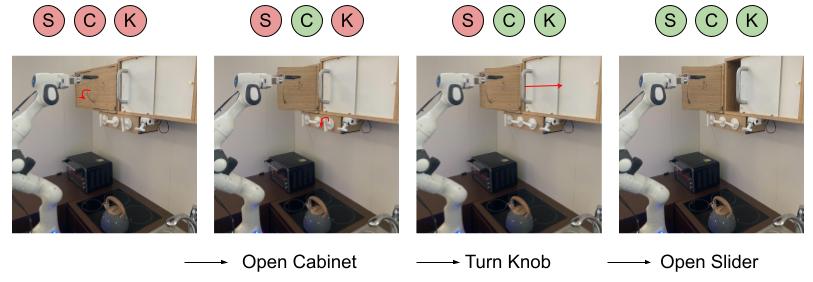}
    \caption{\footnotesize{Multi-step behavior in the kitchen that transitions between having all elements closed to all open, by first having the graph search command the agent to open the cabinet, then turn the knob and open the slider.}}
    \label{fig:filmstrip}
\vspace{-0.4cm}
\end{figure*}



We start by evaluating the ability of DBAP to learn multiple long horizon goal-reaching tasks in the real world. We trained the robot for over $25$ hours as described in Section~\ref{sec:method}, only requiring a reset every $50$ trajectories (in wall-clock time corresponding to once an hour only). This makes training much more practical in terms of human effort. 

As seen from the evaluation performance of offline RL on long horizon success in Table~\ref{tab:success-rates-goalreaching-realworld},
DBAP starts off doing reasonably well during pre-training, achieving a $83\%$ success rate, but improves significantly during autonomous practicing (with only very occasional resets) to around a $95\%$ success rate, indicating the importance of online finetuning. In Fig.~\ref{fig:filmstrip}, we show a qualitative example of DBAP succeeding at one of the 8 long horizon tasks, where the robot must reach a goal with slider, cabinet, and microwave door all open, from a state where all three are closed. The graph search automatically commands the low-level policy to reach each of the subgoals for the complete task. The learned behaviors are best appreciated by viewing the supplementary video.

In comparison, the no-pretraining, from-scratch baseline results in 0\% success rate. This indicates the importance of being able to incorporate prior data, as is done in DBAP, to overcome hard exploration challenges inherent in long horizon manipulation. DBAP also significantly outperforms imitation learning in terms of long horizon goal-reaching performance. We compare the number of tasks commanded to accomplish the multi-step goals and observe the average path length is significantly lower for our method than baselines. 


\begin{figure*}[!ht]
    \centering
        \includegraphics[width=0.4\textwidth]{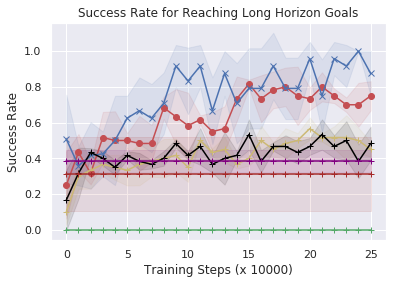}
    \includegraphics[width=0.4\textwidth]{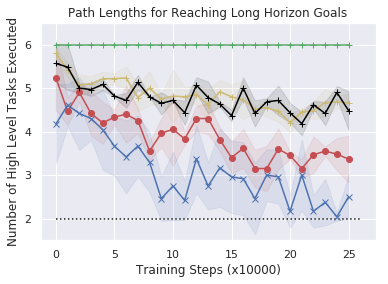} \\
    \includegraphics[width = 0.8\textwidth]{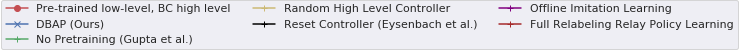}
    \caption{\footnotesize{Performance on the simulated environment in Fig~\ref{fig:sim_env}
    \textbf{(Left)} Success rate averaged across 3 seeds each on long horizon goals \textbf{(Right)} Average number of sub-goals commanded to reach a long horizon goal}}
    \label{fig:sim_results_goalreaching}
    \vspace{-0.5cm}
\end{figure*}

\subsection{Demonstration-Bootstrapped Practicing in Simulation}

Next, to study comparisons with baselines more systematically we evaluate the ability of DBAP to learn to accomplish multi-step goals in simulation as described in Section~\ref{sec:system}, 
with a minimal amounts of automatically provided resets. In simulation we assume access to one human reset every $10$ episodes, reducing the amount of human supervision by an order of magnitude as compared to standard RL.
As shown in Fig~\ref{fig:sim_results_goalreaching} (left), our method  successfully reaches a variety of multi-step goals and shows improvement over time with minimal human involvement (from $50\%$  to $90\%$ success rate). While other techniques are able to occasionally succeed, they take significantly longer to get to the goal, often resulting in roundabout paths to achieve the goal. This leads to the extended path length of trajectories in Fig~\ref{fig:sim_results_goalreaching} (middle). Training a high level policy via goal relabeling and BC (the ``pretrained low-level, BC task selection'' and ``Full relabeled relay policy learning" baseline in Fig~\ref{fig:sim_results_goalreaching}.) can inherit the biases of the data. If the play data demonstrates sub-optimal paths to reach long horizon goals, this is reflected in the performance of the high-level BC. This also happens for the offline imitation learning baseline, indicating it is not a facet of offline reinforcement learning. In contrast, our graph-based search method is able to find and practice shorter paths than the baselines. Importantly, this improvement is acquired with a minimal requirement for human effort in the process.  

These studies suggest that DBAP is able to both enable autonomous practicing with minimal human effort \emph{and} via this practicing enable a real world robotic agent to solve long horizon tasks by sequencing subgoals appropriately.

\subsection{Visualizations and Ablations}
In this section, we perform a number of visualizations and ablations to understand the behavior of the algorithm and the human provided data. In particular we conduct experiments to analyze (1) the comparison between graph search and behavior cloned high level policies to understand where the benefits come from (2) the effect of the entropy maximizing objective while choosing goals to practice on, as compared to a random walk (3) effect of reset frequency on performance and finally (4) we visualize the nature of the human provided demonstration data. 

\begin{figure}[!h]
\begin{minipage}{0.47\textwidth}
    \centering
    \includegraphics[width=0.7\textwidth]{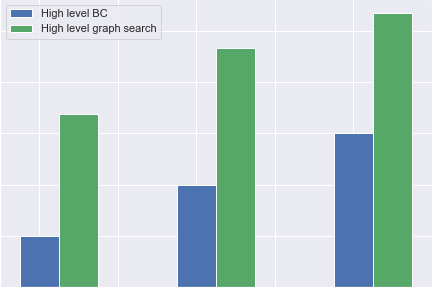}
    \caption{Comparison of path length of BC vs graph search for a simulated problem assuming perfect low level. We see that high level BC can often be ineffective at learning short paths to a goal}
    \label{fig:ablation}
\end{minipage}
\hspace{0.3cm}
\begin{minipage}{0.47\textwidth}
    \centering
    \includegraphics[width=0.9\textwidth]{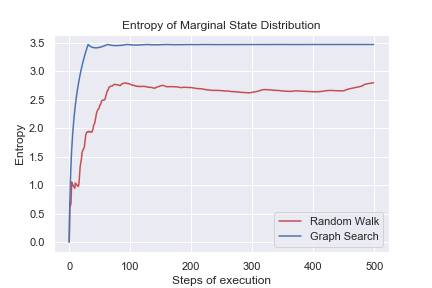}
    \caption{\footnotesize{Visualization of entropy on a 32 state chain MDP. We see that the  entropy of the marginal distribution over states is much higher with graph search than a random walk.}}
    \label{fig:chainmdp_viz}
\end{minipage}
\end{figure}

We fist visualize the sequence of tasks proposed by our graph search algorithm during evaluation time as compared to the behavior cloning baseline mentioned above. The behavior cloning baseline trains the high level with goal conditioned behavior cloning while graph search simply builds a graph of feasible edges and performs search. We find that when we run an idealized experiment, where the multi-task policies are assumed to be perfect, the effective path length obtained by BC is significantly higher than graph search. This suggests that doing high level learning with relabeled BC as suggested in ~\citep{gupta2019relay, lynch2020learning} is prone to an issue of cycles where it is not trained to take the shortest path if the training data is cyclic (as is common in play data). Graph search on the other hand, avoids these issues and is able to find a shortest path to the goal easily as seen in Fig~\ref{fig:ablation}.

To further understand the behavior of graph search based practicing autonomously, we then ran an isolated analysis experiment on a very simple chain MDP environment to understand the performance of the graph search algorithm in terms of task visitations and entropy over the distribution of tasks, as compared to simply performing a random walk on tasks. We visualize these results in Fig~\ref{fig:chainmdp_viz}, where entropy of the marginal distribution of states of interest is plotted against steps of the training process. We find that while both have increasing entropy, graph based search has much higher entropy as it maintains a uniform likelihood over states, providing the coverage needed to achieve good performance in the evaluations in Fig~\ref{fig:chainmdp_viz}.

\begin{figure}[!h]
    \centering
    \vspace{-0.1cm}
    \includegraphics[width=0.24\textwidth]{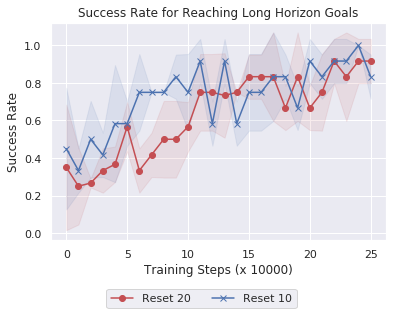}
    \includegraphics[width=0.24\textwidth]{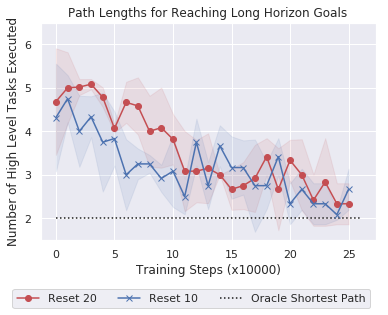}
    \vspace{-0.1cm}
    \caption{\footnotesize{Visualization of the effect of reset frequency on long horizon performance.}}
    \label{fig:reset_ablation}
\end{figure}

We next measured the effect of reset frequency by running an experiment where we vary the number of resets $n$. We see that there is no significant degradation from 10 to 20 resets but this is likely to degrade significantly as increased. This is shown in Fig~\ref{fig:reset_ablation}.

\begin{figure}[!h]
\begin{minipage}{0.23\textwidth}
    \centering
    \includegraphics[width=\textwidth]{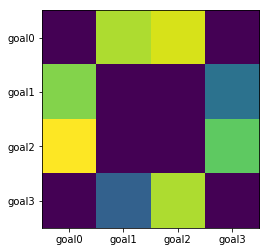}
    \vspace{-0.5cm}
    \caption{Visualization of distribution of demonstration data provided in simulation. The heatmap denotes proportion of demonstrations transitioning between each of the $4$ goals of interest. }
    \label{fig:demo_distribution_sim}
\end{minipage}
\begin{minipage}{0.23\textwidth}
    \centering
    \includegraphics[width=\textwidth]{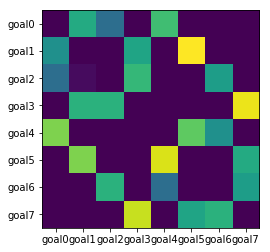}
    \vspace{-0.6cm}
    \caption{Visualization of distribution of demonstration data provided in the real world. The heatmap denotes proportion of demonstrations transitioning between each of the $8$ goals of interest.}
    \label{fig:demo_distribution_hardware}
\end{minipage}
\end{figure}

Lastly, we visualize the quality of the human provided demonstration data. We plot a heatmap over different transitions that were provided in the human demonstrated data, showing the coverage over states in Fig~\ref{fig:demo_distribution_hardware}, Fig~\ref{fig:demo_distribution_sim}.

These ablation experiments yield further insight into the various elements of our proposed system and show that the proposed graph search algorithm yields benefits in term of state coverage. 

%% file: texs/discussion.tex
\vspace{-0.2cm}
\section{Discussion}
\label{sec:discussion}

In this work, we described the design principles behind a system for learning behaviors autonomously using RL. We leverage human-provided data to bootstrap a multi-task RL system, using some tasks to provide resets for others. Our method combines offline RL at the low level and model-based graph search at the high level, where prior data is used to bootstrap low-level policies as well as facilitate practicing. We demonstrated this on simulated and real-world kitchen environments.

This work has several imitations. It requires tasks to be discretely defined, which is done by thresholding low-level states. Extending to high dimensional problems is an interesting avenue for future work. The algorithm assumes that the environment does not transition into ``irreversible" states. This assumption is common in reset-free RL ~\cite{gupta2021reset, zhu2020ingredients, eysenbach2017leave}, and encompasses a range of robotics problems. Expanding the system to cope with scenarios where the kettle falls off the table or the environment goes into irrecoverable states would be an interesting future direction.

There are a number of interesting future directions we can pursue based on this work. One would be to obviate the need for state estimation, instead learning directly from visual inputs. Additionally, using these visual inputs to infer rewards directly will be important going forward. Moreover, expanding the system to a mobile manipulator to deal with scenarios where the kettle falls off the table or the environment goes into irrecoverable states would be important. For building unconstrained reset free settings going forward, allowing for constantly expanding graphs where new tasks of interest are constantly being added and incorporated would be interesting to explore. In doing so, we can build systems where every time the robot gets stuck, new tasks are incrementally added to provide resets, with the system incrementally getting more and more autonomous over time. 

%% file: texs/acknowledgements.tex
\section{Acknowledgements}
This authors would like to thank several colleagues at Robotics at Google and UC Berkeley, specifically Thinh Nguyen, Gus Kouretas, Jeff Seto, Krista Reymann, Michael Ahn, Ken Oslund and Sherry Moore for their contributions with building the hardware setup and Archit Sharma, Kelvin Xu, Justin Yu, Vikash Kumar for helpful discussions. The authors would also like to acknowledge funding from the Office of Naval Research (ONR) for SL and AG. 

%% file: texs/appendix.tex
\section{Appendix}

\subsection{Hyperparameters and Model Architectures}
Here we list a set of general hyperparameters and details of the model training. If indiciated as a list, we performed a grid search over those values and the underlined value is the chosen one to report in the paper. 

\begin{table}[!h]
    \centering
    \begin{tabular}{| p{5cm}||p{4cm} |}
     \hline
     \textbf{AWAC} & \\
     \hline
     Learning Rate & $3 \times 10^{-4}$\\
     Discount Factor $\gamma$ & $0.99$\\
     Policy Type & Gaussian\\
     Policy Hidden Sizes & $(256, 256)$\\
     Policy Hidden Activation & ReLU\\
     Policy Weight Decay & $10^{-4}$ \\
     Policy Learning Rate & $3\times10^{-4}$ \\
     RL Batch Size & $1024$ \\
     Reward Scaling & $1$\\
     Replay Buffer Size & $500,000$\\
     Q Hidden Sizes & $(512, 512)$\\
     Q Hidden Activation & ReLU\\
     Q Weight Decay & $0$ \\
     Q Learning Rate & $3 \times 10^{-4}$\\
     Target Network $\tau$ & $5\times10^{-3}$ \\
     Temperature ($\beta$) & $30.$ \\
     Epoch Length & $2000$ \\
     Path Length & $200$ \\
     Max High Level Steps & $6$ \\
     Standard Deviation Lower Bound & $10^{-5}$ \\
     Pretraining Steps & $30000$ \\
     \hline 
     \textbf{Behavior Cloning} & \\
     \hline
     Hidden Layer Sizes & $(256, 256)$\\
     Training Steps & $30000$\\
     \hline
    \end{tabular}
\caption{Hyperparameters used for experiments}
\end{table}

As mentioned above, we use a standard 2 hidden layer MLP to represent policies and value functions. The state space consists of the position of the end-effector, euler rotation of the end effector, position of the various elements in the scene and a representation of the goal. In the hardware experiments, the representation of the goals of interest is a continuous vector of the position of the end-effector, euler rotation of the end effector, position of the various elements in the scene for a particular goal of interest. In the simulation environments, this is a discrete one-hot vector representing the ID of the particular goal of interest being commanded. Beyond the experiments mentioned in the main paper, we also ran experiments resetting every $50$ episodes instead of 10, and achieved $96\%$ success rate as well. 

\subsection{Reward Functions}

We used a generic and simple reward function for different goals $s_g$, as follows:

\begin{align}
    r(s, a, s_g) = -20*\|x_{\text{ee}} - x_{\text{element}}\|_2 -20*\|\theta_{\text{element}} - \theta_{\text{goal}}\|_2
\end{align}

Here $x_{\text{ee}}$ corresponds to the position of the end effector, $x_{\text{element}}$ corresponds to the position of the particular element (cabinet, slider, knob) that has to be moved to accomplish the particular goal $s_g$,  $\theta_{\text{element}}$ is the current position of the above-mentioned element and $\theta_{\text{goal}}$ is the goal position of the element being manipulated. This reward function essentially encourages the arm to move towards the element of interest and then manipulate it towards it's goal position. This reward function is suitable for articulated objects, but may have to be replaced by a more involved reward function for scenes with more free objects. 

\subsection{Baseline Details}
We provide some further details of the baseline methods below: 

\textbf{Non-pretrained, graph search task selection}
This baseline simply uses the exact same graph search algorithm at the high level but starts the low level policy completely from scratch and trains the algorithm exactly the same way as \method, using exactly the same hyperparameters. 

\textbf{Pretrained low-level, random high level controller}
This baseline uses completely random goal selection both during practicing and long horizon task accomplishment. It simply samples a goal of interest from the set of all possible goals of interest 

\textbf{Pretrained low-level, BC task selection}
This baseline uses a behavior cloned high level model to select goals, where the high level task selector $q(s_g|s, s_g^{\text{desired}})$ is trained using behavior cloning on human provided data, and goals are selected using this task selector. In particular, we relabel sequences of goals of interest visited in the data within a particular window (as outlined in ~\cite{gupta2019relay}) to generate $(s, s_g, s_g^{\text{desired}})$ which can then be used to train $q(s_g|s, s_g^{\text{desired}})$ via supervised learning. During autonomous practicing, goals are chosen by choosing a $s_g^{\text{desired}}$ at random and then sampling a next goal from the behavior cloned $q(s_g|s, s_g^{\text{desired}})$ conditioned on the chosen $s_g^{\text{desired}}$. At long horizon execution time, the agent simply sets $s_g^{\text{desired}}$ to the desired long horizon goal and sample from $q(s_g|s, s_g^{\text{desired}})$. 

\textbf{Pretrained low-level, reset controller}
This baseline alternates between setting the first goal of interest as the goal and sampling a goal of interest at random and setting this at the goal at the high level. 

All of these baselines were implemented using exactly the same underlying algorithm and the same hyperparameters as mentioned above.

\subsection{Details on Task Thresholding}
Rather than performing graph search on the infinite space of continuous states, a graph is built over the discrete states of interest in the environment. At the end of each episode the current state $s$ is discretized in order to actually find the closest goal of interest and this is then used to choose the next state to command. The procedure used for discretization is very simple - it defines an open/close state for every articulated element in the scene (not the end effector) and then the discrete state is determined by thresholding every element to either be open or closed and then converting the binary representation of all the open(1) or closed(0) elements into an integer. This allows the states to be discretized in a way that it is easy for graph search to operate.





%% file: main.bbl
\begin{thebibliography}{40}
\providecommand{\natexlab}[1]{#1}
\providecommand{\url}[1]{\texttt{#1}}
\expandafter\ifx\csname urlstyle\endcsname\relax
  \providecommand{\doi}[1]{doi: #1}\else
  \providecommand{\doi}{doi: \begingroup \urlstyle{rm}\Url}\fi

\bibitem[Bacon et~al.(2017)Bacon, Harb, and Precup]{plb2017optioncritic}
Pierre{-}Luc Bacon, Jean Harb, and Doina Precup.
\newblock The option-critic architecture.
\newblock In \emph{AAAI}, 2017.

\bibitem[Barto and Mahadevan(2003)]{barto2003hrl}
Andrew~G. Barto and Sridhar Mahadevan.
\newblock Recent advances in hierarchical reinforcement learning.
\newblock \emph{Discret. Event Dyn. Syst.}, 2003.

\bibitem[Billard et~al.(2008)Billard, Calinon, Dillmann, and
  Schaal]{billard2008survey}
Aude Billard, Sylvain Calinon, Ruediger Dillmann, and Stefan Schaal.
\newblock Survey: Robot programming by demonstration.
\newblock Technical report, Springrer, 2008.

\bibitem[Co-Reyes et~al.(2020)Co-Reyes, Sanjeev, Berseth, Gupta, and
  Levine]{co2020ecological}
John~D Co-Reyes, Suvansh Sanjeev, Glen Berseth, Abhishek Gupta, and Sergey
  Levine.
\newblock Ecological reinforcement learning.
\newblock \emph{arXiv preprint arXiv:2006.12478}, 2020.

\bibitem[Dayan and Hinton(1992)]{dayan92feudal}
Peter Dayan and Geoffrey~E. Hinton.
\newblock Feudal reinforcement learning.
\newblock In \emph{NeurIPS}, 1992.

\bibitem[Dietterich(1998)]{dietterich98maxq}
Thomas~G. Dietterich.
\newblock The {MAXQ} method for hierarchical reinforcement learning.
\newblock In \emph{ICML}, 1998.

\bibitem[Dijkstra(1959)]{dijkstra1959note}
Edsger~W Dijkstra.
\newblock A note on two problems in connexion with graphs.
\newblock \emph{Numerische mathematik}, 1\penalty0 (1):\penalty0 269--271,
  1959.

\bibitem[Eysenbach et~al.(2019)Eysenbach, Salakhutdinov, and
  Levine]{eysenbach2019sorb}
Ben Eysenbach, Ruslan Salakhutdinov, and Sergey Levine.
\newblock Search on the replay buffer: Bridging planning and reinforcement
  learning.
\newblock In Hanna~M. Wallach, Hugo Larochelle, Alina Beygelzimer, Florence
  d'Alch{\'{e}}{-}Buc, Emily~B. Fox, and Roman Garnett, editors,
  \emph{NeurIPS}, 2019.

\bibitem[Eysenbach et~al.(2017)Eysenbach, Gu, Ibarz, and
  Levine]{eysenbach2017leave}
Benjamin Eysenbach, Shixiang Gu, Julian Ibarz, and Sergey Levine.
\newblock Leave no trace: Learning to reset for safe and autonomous
  reinforcement learning.
\newblock \emph{arXiv preprint arXiv:1711.06782}, 2017.

\bibitem[Fujimoto et~al.(2019)Fujimoto, Meger, and Precup]{fujimoto2019off}
Scott Fujimoto, David Meger, and Doina Precup.
\newblock Off-policy deep reinforcement learning without exploration.
\newblock In \emph{International Conference on Machine Learning}, pages
  2052--2062. PMLR, 2019.

\bibitem[Gao et~al.(2018)Gao, Xu, Lin, Yu, Levine, and
  Darrell]{gao2018reinforcement}
Yang Gao, Huazhe Xu, Ji~Lin, Fisher Yu, Sergey Levine, and Trevor Darrell.
\newblock Reinforcement learning from imperfect demonstrations.
\newblock \emph{arXiv preprint arXiv:1802.05313}, 2018.

\bibitem[Garcia et~al.(1989)Garcia, Prett, and Morari]{garcia89surveympc}
Carlos~E. Garcia, David~M. Prett, and Manfred Morari.
\newblock Model predictive control: Theory and practice - {A} survey.
\newblock \emph{Autom.}, 25\penalty0 (3):\penalty0 335--348, 1989.
\newblock \doi{10.1016/0005-1098(89)90002-2}.
\newblock URL \url{https://doi.org/10.1016/0005-1098(89)90002-2}.

\bibitem[Gupta et~al.(2019)Gupta, Kumar, Lynch, Levine, and
  Hausman]{gupta2019relay}
Abhishek Gupta, Vikash Kumar, Corey Lynch, Sergey Levine, and Karol Hausman.
\newblock Relay policy learning: Solving long horizon tasks via imitation and
  reinforcement learning.
\newblock \emph{Conference on Robot Learning (CoRL)}, 2019.

\bibitem[Gupta et~al.(2021)Gupta, Yu, Zhao, Kumar, Rovinsky, Xu, Devlin, and
  Levine]{gupta2021reset}
Abhishek Gupta, Justin Yu, Tony~Z Zhao, Vikash Kumar, Aaron Rovinsky, Kelvin
  Xu, Thomas Devlin, and Sergey Levine.
\newblock Reset-free reinforcement learning via multi-task learning: Learning
  dexterous manipulation behaviors without human intervention.
\newblock \emph{arXiv preprint arXiv:2104.11203}, 2021.

\bibitem[Ha et~al.(2020)Ha, Xu, Tan, Levine, and Tan]{ha2020learning}
Sehoon Ha, Peng Xu, Zhenyu Tan, Sergey Levine, and Jie Tan.
\newblock Learning to walk in the real world with minimal human effort.
\newblock \emph{arXiv preprint arXiv:2002.08550}, 2020.

\bibitem[Han et~al.(2015)Han, Levine, and Abbeel]{han2015learning}
Weiqiao Han, Sergey Levine, and Pieter Abbeel.
\newblock Learning compound multi-step controllers under unknown dynamics.
\newblock In \emph{2015 IEEE/RSJ International Conference on Intelligent Robots
  and Systems (IROS)}, pages 6435--6442. IEEE, 2015.

\bibitem[Harutyunyan et~al.(2019)Harutyunyan, Dabney, Mesnard, Azar, Piot,
  Heess, van Hasselt, Wayne, Singh, Precup, and Munos]{harutyunyan2019hca}
Anna Harutyunyan, Will Dabney, Thomas Mesnard, Mohammad~Gheshlaghi Azar, Bilal
  Piot, Nicolas Heess, Hado van Hasselt, Gregory Wayne, Satinder Singh, Doina
  Precup, and R{\'{e}}mi Munos.
\newblock Hindsight credit assignment.
\newblock In \emph{NeurIPS}, 2019.

\bibitem[Kostrikov et~al.(2021)Kostrikov, Tompson, Fergus, and
  Nachum]{kostrikov2021offline}
Ilya Kostrikov, Jonathan Tompson, Rob Fergus, and Ofir Nachum.
\newblock Offline reinforcement learning with fisher divergence critic
  regularization.
\newblock \emph{arXiv preprint arXiv:2103.08050}, 2021.

\bibitem[Kreidieh et~al.(2019)Kreidieh, Parajuli, Lichtle, You, Nasr, and
  Bayen]{kreidieh2019marlhrl}
Abdul~Rahman Kreidieh, Samyak Parajuli, Nathan Lichtle, Yiling You, Rayyan
  Nasr, and Alexandre~M. Bayen.
\newblock Inter-level cooperation in hierarchical reinforcement learning.
\newblock \emph{arXiv}, 2019.

\bibitem[Kumar et~al.(2019)Kumar, Fu, Tucker, and Levine]{kumar2019stabilizing}
Aviral Kumar, Justin Fu, George Tucker, and Sergey Levine.
\newblock Stabilizing off-policy q-learning via bootstrapping error reduction.
\newblock \emph{arXiv preprint arXiv:1906.00949}, 2019.

\bibitem[Kumar et~al.(2020)Kumar, Zhou, Tucker, and
  Levine]{kumar2020conservative}
Aviral Kumar, Aurick Zhou, George Tucker, and Sergey Levine.
\newblock Conservative q-learning for offline reinforcement learning.
\newblock \emph{arXiv preprint arXiv:2006.04779}, 2020.

\bibitem[Le et~al.(2018)Le, Jiang, Agarwal, Dud{\'{\i}}k, Yue, and
  III]{le2018hirl}
Hoang~Minh Le, Nan Jiang, Alekh Agarwal, Miroslav Dud{\'{\i}}k, Yisong Yue, and
  Hal~Daum{\'{e}} III.
\newblock Hierarchical imitation and reinforcement learning.
\newblock In Jennifer~G. Dy and Andreas Krause, editors, \emph{ICML}, 2018.

\bibitem[Levine et~al.(2020)Levine, Kumar, Tucker, and Fu]{levine2020offline}
Sergey Levine, Aviral Kumar, George Tucker, and Justin Fu.
\newblock Offline reinforcement learning: Tutorial, review, and perspectives on
  open problems.
\newblock \emph{arXiv preprint arXiv:2005.01643}, 2020.

\bibitem[Levy et~al.(2019)Levy, Konidaris, Jr., and Saenko]{levy2019hac}
Andrew Levy, George~Dimitri Konidaris, Robert~Platt Jr., and Kate Saenko.
\newblock Learning multi-level hierarchies with hindsight.
\newblock In \emph{ICLR}, 2019.

\bibitem[Lynch et~al.(2020)Lynch, Khansari, Xiao, Kumar, Tompson, Levine, and
  Sermanet]{lynch2020learning}
Corey Lynch, Mohi Khansari, Ted Xiao, Vikash Kumar, Jonathan Tompson, Sergey
  Levine, and Pierre Sermanet.
\newblock Learning latent plans from play.
\newblock In \emph{Conference on Robot Learning}, pages 1113--1132. PMLR, 2020.

\bibitem[Mandlekar et~al.(2020{\natexlab{a}})Mandlekar, Ramos, Boots, Savarese,
  Li, Garg, and Fox]{mandlekar2020iris}
Ajay Mandlekar, Fabio Ramos, Byron Boots, Silvio Savarese, Fei{-}Fei Li,
  Animesh Garg, and Dieter Fox.
\newblock {IRIS:} implicit reinforcement without interaction at scale for
  learning control from offline robot manipulation data.
\newblock In \emph{2020 {IEEE} International Conference on Robotics and
  Automation, {ICRA} 2020, Paris, France, May 31 - August 31, 2020}, pages
  4414--4420. {IEEE}, 2020{\natexlab{a}}.
\newblock \doi{10.1109/ICRA40945.2020.9196935}.
\newblock URL \url{https://doi.org/10.1109/ICRA40945.2020.9196935}.

\bibitem[Mandlekar et~al.(2020{\natexlab{b}})Mandlekar, Xu,
  Mart{\'{\i}}n{-}Mart{\'{\i}}n, Savarese, and Fei{-}Fei]{mandlekar2020gti}
Ajay Mandlekar, Danfei Xu, Roberto Mart{\'{\i}}n{-}Mart{\'{\i}}n, Silvio
  Savarese, and Li~Fei{-}Fei.
\newblock {GTI:} learning to generalize across long-horizon tasks from human
  demonstrations.
\newblock In Marc Toussaint, Antonio Bicchi, and Tucker Hermans, editors,
  \emph{Robotics: Science and Systems XVI, Virtual Event / Corvalis, Oregon,
  USA, July 12-16, 2020}, 2020{\natexlab{b}}.
\newblock \doi{10.15607/RSS.2020.XVI.061}.
\newblock URL \url{https://doi.org/10.15607/RSS.2020.XVI.061}.

\bibitem[Manschitz et~al.(2015)Manschitz, Kober, Gienger, and
  Peters]{manschitz2015learning}
Simon Manschitz, Jens Kober, Michael Gienger, and Jan Peters.
\newblock Learning movement primitive attractor goals and sequential skills
  from kinesthetic demonstrations.
\newblock \emph{Robotics and Autonomous Systems}, 74:\penalty0 97--107, 2015.

\bibitem[Matsushima et~al.(2020)Matsushima, Furuta, Matsuo, Nachum, and
  Gu]{deploymentorl}
Tatsuya Matsushima, Hiroki Furuta, Yutaka Matsuo, Ofir Nachum, and Shixiang Gu.
\newblock Deployment-efficient reinforcement learning via model-based offline
  optimization.
\newblock \emph{CoRR}, abs/2006.03647, 2020.
\newblock URL \url{https://arxiv.org/abs/2006.03647}.

\bibitem[M{\"u}lling et~al.(2013)M{\"u}lling, Kober, Kroemer, and
  Peters]{mulling2013learning}
Katharina M{\"u}lling, Jens Kober, Oliver Kroemer, and Jan Peters.
\newblock Learning to select and generalize striking movements in robot table
  tennis.
\newblock \emph{The International Journal of Robotics Research}, 32\penalty0
  (3):\penalty0 263--279, 2013.

\bibitem[Nachum et~al.(2018)Nachum, Gu, Lee, and Levine]{nachum18hiro}
Ofir Nachum, Shixiang Gu, Honglak Lee, and Sergey Levine.
\newblock Data-efficient hierarchical reinforcement learning.
\newblock In \emph{NeurIPS}, 2018.

\bibitem[Nair et~al.(2020)Nair, Dalal, Gupta, and Levine]{nair2020accelerating}
Ashvin Nair, Murtaza Dalal, Abhishek Gupta, and Sergey Levine.
\newblock Accelerating online reinforcement learning with offline datasets.
\newblock \emph{arXiv preprint arXiv:2006.09359}, 2020.

\bibitem[Peng et~al.(2019)Peng, Kumar, Zhang, and Levine]{peng2019advantage}
Xue~Bin Peng, Aviral Kumar, Grace Zhang, and Sergey Levine.
\newblock Advantage-weighted regression: Simple and scalable off-policy
  reinforcement learning.
\newblock \emph{arXiv preprint arXiv:1910.00177}, 2019.

\bibitem[Rajeswaran et~al.(2017)Rajeswaran, Kumar, Gupta, Vezzani, Schulman,
  Todorov, and Levine]{rajeswaran2017learning}
Aravind Rajeswaran, Vikash Kumar, Abhishek Gupta, Giulia Vezzani, John
  Schulman, Emanuel Todorov, and Sergey Levine.
\newblock Learning complex dexterous manipulation with deep reinforcement
  learning and demonstrations.
\newblock \emph{arXiv preprint arXiv:1709.10087}, 2017.

\bibitem[Savinov et~al.(2018)Savinov, Dosovitskiy, and Koltun]{savinov18sptm}
Nikolay Savinov, Alexey Dosovitskiy, and Vladlen Koltun.
\newblock Semi-parametric topological memory for navigation.
\newblock In \emph{ICLR}, 2018.

\bibitem[Singh et~al.(2020)Singh, Yu, Yang, Zhang, Kumar, and
  Levine]{singh2020cog}
Avi Singh, Albert Yu, Jonathan Yang, Jesse Zhang, Aviral Kumar, and Sergey
  Levine.
\newblock Cog: Connecting new skills to past experience with offline
  reinforcement learning.
\newblock \emph{arXiv preprint arXiv:2010.14500}, 2020.

\bibitem[Sutton et~al.(1999)Sutton, Precup, and Singh]{sutton99options}
Richard~S. Sutton, Doina Precup, and Satinder~P. Singh.
\newblock Between mdps and semi-mdps: {A} framework for temporal abstraction in
  reinforcement learning.
\newblock \emph{Artif. Intell.}, 1999.

\bibitem[Wang et~al.(2020)Wang, Novikov, {\.Z}o{\l}na, Springenberg, Reed,
  Shahriari, Siegel, Merel, Gulcehre, Heess, et~al.]{wang2020critic}
Ziyu Wang, Alexander Novikov, Konrad {\.Z}o{\l}na, Jost~Tobias Springenberg,
  Scott Reed, Bobak Shahriari, Noah Siegel, Josh Merel, Caglar Gulcehre,
  Nicolas Heess, et~al.
\newblock Critic regularized regression.
\newblock \emph{arXiv preprint arXiv:2006.15134}, 2020.

\bibitem[Yu et~al.(2021)Yu, Kumar, Rafailov, Rajeswaran, Levine, and
  Finn]{yu2021combo}
Tianhe Yu, Aviral Kumar, Rafael Rafailov, Aravind Rajeswaran, Sergey Levine,
  and Chelsea Finn.
\newblock Combo: Conservative offline model-based policy optimization.
\newblock \emph{arXiv preprint arXiv:2102.08363}, 2021.

\bibitem[Zhu et~al.(2020)Zhu, Yu, Gupta, Shah, Hartikainen, Singh, Kumar, and
  Levine]{zhu2020ingredients}
Henry Zhu, Justin Yu, Abhishek Gupta, Dhruv Shah, Kristian Hartikainen, Avi
  Singh, Vikash Kumar, and Sergey Levine.
\newblock The ingredients of real-world robotic reinforcement learning.
\newblock \emph{arXiv preprint arXiv:2004.12570}, 2020.

\end{thebibliography}
